\documentclass[11pt,a4paper]{article}
\usepackage[hyperref]{aacl-ijcnlp2020}
\usepackage{times}
\usepackage{latexsym}
\usepackage{graphicx}
\usepackage{amsmath}
\usepackage{float}
\usepackage{hyperref}
\usepackage{microtype}

\aclfinalcopy % Uncomment this line for the final submission
\def\aclpaperid{345} %  Enter the aacl-ijcnlp Paper ID here

\renewcommand\vec[1]{\overrightarrow{#1}}
\newcommand\cev[1]{\overleftarrow{#1}}

\title{Systematic Generalization on gSCAN \\with Language Conditioned Embedding}

\author{
  Tong Gao\thanks{(*) denotes co-first authorship, authors contribute equally and are listed in alphabetical order.} \hfill Qi Huang\footnotemark[1] \hfill Raymond J. Mooney\\
  Department of Computer Science\\
  University of Texas at Austin\\
  \texttt{\{gaotong,qhuang,mooney\}@cs.utexas.edu}
  }

\date{}

\begin{document}
\maketitle
\begin{abstract}
Systematic Generalization refers to a learning algorithm's ability to extrapolate learned behavior to unseen situations that are distinct but semantically similar to its training data. As shown in recent work, state-of-the-art deep learning models fail dramatically even on tasks for which they are designed when the test set is systematically different from the training data. We hypothesize that explicitly modeling the relations between objects in their contexts while learning their representations will help achieve systematic generalization. Therefore, we propose a novel method that learns objects' contextualized embedding with dynamic message passing conditioned on the input natural language and is end-to-end trainable with other downstream deep learning modules. To our knowledge, this model is the first one that significantly outperforms the provided baseline and reaches state-of-the-art performance on \emph{grounded} SCAN (gSCAN), a grounded natural language navigation dataset designed to require systematic generalization in its test splits.
\end{abstract}

\section{Introduction}
Systematic Generalization refers to a learning algorithm's ability to extrapolate learned behavior to unseen situations that are distinct but semantically similar to its training data. It has long been recognized as a key aspect of humans' cognitive capacities \citep{connectionism}. Specifically, humans' mastery of systematic generalization is prevalent in grounded natural language understanding. For example, humans can reason about the relations between all pairs of concepts from two domains, even if they have only seen a small subset of pairs during training. If a child observes "red squares", "green squares" and "yellow circles", he or she can recognize "red circles" at their first encounter. Humans can also contextualize their reasoning about objects' attributes. For example, a city being referred to as "the larger one" within a state might be referred to as "the smaller one" nationwide. In the past decade, deep neural networks have shown tremendous success on a collection of grounded natural language processing tasks, such as visual question answering (VQA), image captioning, and vision-and-language navigation \citep{mac,bottomup,navigation}. Despite all the success, recent literature shows that current deep learning approaches are exploiting statistical patterns discovered in the datasets to achieve high performance, an approach that does not achieve systematic generalization. \citet{gururang} discovered that annotation artifacts like negation words or purpose clauses in natural language inference data can be used by simple text classification categorization model to solve the given task. \citet{jia2017} demonstrated that adversarial examples can fool reading comprehension systems. Indeed, deep learning models often fail to achieve systematic generalizations even on tasks on which they are claimed to perform well. As shown by \citet{systematic}, state-of-the-art Visual Questioning Answering (VQA) \citep{mac,film} models fail dramatically even on a synthetic VQA dataset designed with systematic difference between training and test sets.

In this work, we focus on approaching systematic generalization in grounded natural language understanding tasks. We experiment with a recently introduced synthetic dataset, \emph{grounded} SCAN (gSCAN), that requires systematic generalization to solve \citep{ruis2020benchmark}. For example, after observing how to "walk hesitantly" to a target object in a grid world, the learning agent is tested with instruction that requires it to "pull hesitantly", therefore testing its ability to generalize adverbs to unseen adverb-verb combinations.

When presented with a world of objects with different attributes, and natural language sentences that describe such objects, the goal of the model is to generalize its ability to understand unseen sentences describing novel combinations of observed objects, or even novel objects with observed attributes. One of the essential steps in achieving this goal is to obtain good object embeddings to which natural language can be grounded. By considering each object as a bag of its descriptive attributes, this problem is further transformed into learning good representations for those attributes based on the training data. This requires: 1) learning good representations of attributes whose actual meanings are contextualized, for example, "smaller" and "lighter", etc.; 2) learning good representations for attributes so that conceptually similar attributes, e.g., "yellow" and "red", have similar representations. We hypothesize that explicitly modeling the relations between objects in their contexts, i.e., learning contextualized object embeddings, will help achieve systematic generalization. 
This is intuitively helpful for learning concepts with contextualized meaning, just as learning to recognize the "smaller" object in a novel pair requires experience of comparison between semantically similar object pairs. 
Learning contextualized object embeddings can also be helpful for obtaining good representations for semantically similar concepts when such concepts are the only difference between two contexts.
%Qi: for the second argument above (line 76), it's still not well phrased. We kind of trying to summarize intuition from empirical evidence why contextualized object embedding help even in test splits not directly testing on learning context/relation. There is another "idea" that I'm not sure how to put into better words: 1) we can consider modelling relations between objects with different attributes as learning "operators" for attributes that are in the same conceptual category (color -> (yellow, red, blue, ...) In another word, learning those operators actually help "define" this idea of category, just like in Mathematics you define a metric space by defining a metric and a set that such metric operates on.
Inspired by \citet{Hu_2019}, we propose a novel method that learns an object's contextualized embedding with dynamic message passing conditioned on the input natural language. At each round of message passing, our model collects relational information between each object pair, and constructs an object's contextualized embedding as a weighted combination of them. Such weights are dynamically computed conditioned on the input natural sentence. This contextualized object embedding scheme is trained end-to-end with downstream deep modules for specific grounded natural language processing tasks, such as navigation. Experiments show that our approach significantly outperforms a strong baseline on gSCAN.

\section{Related Work}
Research on deep learning models' systematic generalization behavior has gained traction in recent years, with particular focus on natural language processing tasks.

\subsection{Compositionality}
%Qi: I'm not sure whether we need to list the detail of those 5 tests described by the compositionality paper or not.
An idea that is closely related to systematic generalization is compositionality. \citet{partee} define the principle of compositionality as ``The meaning of a whole is a function of the meanings of the parts and of the way they are syntactically combined". \citet{compositionality} synthesizes different interpretations of this abstract principle into 5 theoretically grounded tests to evaluate a model's ability to represent compositionality: 1) Systematicity: if the model can systematically recombine known parts and rules; 2) Productivity: if the model can extend their predictions beyond what they have seen in the training data; 3) Substitutivity; if the model is robust to synonym substitutions; 4) Localism: if the model's composition operations are local or global; and 5) Overgeneralisation: if the model favors rules or exceptions during training. The gSCAN dataset focuses more on capturing the first three tests in a grounded natural language understanding setting, and our proposed model achieves significant performance improvement on test sets relating to systematicity and substitutivity.

\subsection{Systematic Generalization Datasets}

Many systematic generalization datasets have been proposed in recent years  \citep{systematic,chevalier2018babyai,hill2019emergent,lake2017generalization,ruis2020benchmark}. This paper is conceptually most related to the SQOOP dataset proposed by \citet{systematic}, the SCAN dataset proposed by \citet{lake2017generalization}, and the gSCAN dataset proposed by \citet{ruis2020benchmark}.\\
The SQOOP dataset consists of a random number of MNIST-style alphanumeric characters scattered in an image with specific spatial relations ("left", "right", "up", "down") among them \citep{systematic}. The algorithm is tested with a binary decision task of reasoning about whether a specific relation holds between a pair of alphanumeric characters. Systematic difference is created between the testing and training set by only providing supervision on relations for a subset of digit pairs to the learner, while testing its ability to reason about relations between unseen alphanumeric character pairs. For example, the algorithm is tested with questions like ``is S above T" while it never sees a relation involving both S and T during training. Therefore, to fully solve this dataset, it must learn to generalize its understanding of the relation ``above" to unseen pairs of characters. \citet{lake2017generalization} proposed the SCAN dataset and its related benchmark that tests a learning algorithm's ability to perform compositional learning and zero-shot generalization on a natural language command translation task . Given a natural language command with a limited vocabulary, an algorithm needs to translate it into a corresponding action sequence consisting of action tokens from a finite token set. Compared to SQOOP, SCAN tests the algorithm's ability to learn more complicated linguistic generalizations like "walk around left" to "walk around right". SCAN also ensures that the target action sequence is unique, and an oracle solution exists by providing an interpreter function that can unambiguously translate any given command to its target action sequence.

Going beyond SCAN that focuses purely on syntactic aspects of systematic generalization, the gSCAN dataset proposed by \citet{ruis2020benchmark} is an extension of SCAN. It contains a series of systematic generalization tasks that require the learning agent to ground its understanding of natural language commands in a given grid world to produce the correct action token sequence. We choose gSCAN as our benchmark dataset, as its input command sentences are linguistically more complex, and requires processing  multi-modal input.

\subsection{Systematic Generliazation Algorithms}
\citet{systematic} demonstrated that modular networks, with  a carefully chosen module layout, can achieve nearly perfect systematic generalization on SQOOP dataset. Our approach can be considered as a conceptual generalization of theirs. Each object's initial embedding can be considered as a simple affine encoder module, and we learn the connection scheme among these modules conditioned on natural language instead of hand-designing it. \citet{gordon2019permutation} proposed solving the SCAN benchmark by hard-coding their model to be equivariant to all permutations of SCAN's verb primitives. \citet{andreas-2020-good} proposed GECA (``Good-Enough Compositional Augmentation”) that systematically augments the SCAN dataset by identifying sentence fragments with similar syntactic context, and permuting them to generate novel training examples. This line of permutation-invariant approaches is shown to not generalize well on the gSCAN dataset \citep{ruis2020benchmark}. At the time of submission, our method was the first to outperform the strong baseline provided in the gSCAN benchmark, and also the first one to apply language-conditioned message passing to learn contextualized input embedding for systematic generalization tasks. Concurrent to our work, \citet{kuo2020} proposed a family of parse-tree-based compositional RNN networks to enable systematic generalization, and heavily relies on off-the-shelf parsers to produce the network hierarchy. \citet{Deml2020} use an attention-based prediction of the target object's location as an auxilary training task to regularize the model. However, it only improves over the baseline model in \citet{ruis2020benchmark} in a limited subset of test splits. For completeness, we also compare our model's result with the above two concurrent works.

\section{Problem Definition \& Algorithm}
\label{sec:3}
\subsection{Task Definition}
gSCAN contains a series of systematic generalization tasks in a grounded natural language understanding setting. In gSCAN, the learning agent is tested with the task of following a given natural language instruction to navigate in a two-dimensional grid world with objects. This is achieved in the form of generating a sequence of action tokens from a finite action token set $\mathcal{A} = \{walk, push, pull, stay, L_{turn}, R_{turn}\}$ that brings the agent from its starting location to the target location. An object in gSCAN's world state is encoded with a one-hot encoding describing its attributes in three property types: 1) color $\mathcal{C} = \{red, green, blue, yellow\}$ 2) shape $\mathcal{S} = \{circle, square, cylinder\}$ 3) size $\mathcal{D} = \{1, 2, 3, 4\}$. The agent is also encoded as an ``object" in the grid world, with properties including orientation $\mathcal{O} = \{left, right, up, down\}$ and a binary variable $\mathcal{B} = \{yes, no\}$ denoting the presence of the agent. Therefore, the whole grid is represented as a tensor $x^S \in \mathcal{R}^{d \times d \times c}$, where $d$ is the dimension of the grid, and $c = |\mathcal{C}| + |\mathcal{S}| + |\mathcal{D}| + |\mathcal{O}| + |\mathcal{B}|$.  Mathematically, given an input tuple $x = (x^c, x^S)$, where $x^c = \{x^c_1, x^c_2, ..., x^c_n\}$ represents the navigation instruction, the agent needs to predict the correct output action token sequence $y = \{y_1, y_2, ..., y_m\}$. Despite its simple form, this task is quite challenging. For one, generating the correct action token sequence requires understanding the instruction within the context of the agent's current grid world. It also involves connecting specific instructions with complex dynamic patterns. For example, ``pulling" a square will be mapped to a ``pull" command when the square has a size of 1 or 2, but to ``pull pull" when the square has a size of 3 or 4 (a ``heavy" square); ``move cautiously" requires the agent to turn left then turn right  before making the actual move. gSCAN also introduces a series of test sets that have systematic differences from the training set. Computing the correct action token sequences on these test sets requires the model to learn to combine seen concepts into novel combinations, including novel object property combinations, novel contextual references, etc..

\begin{figure*}
	\centering
	\includegraphics[width=\textwidth]{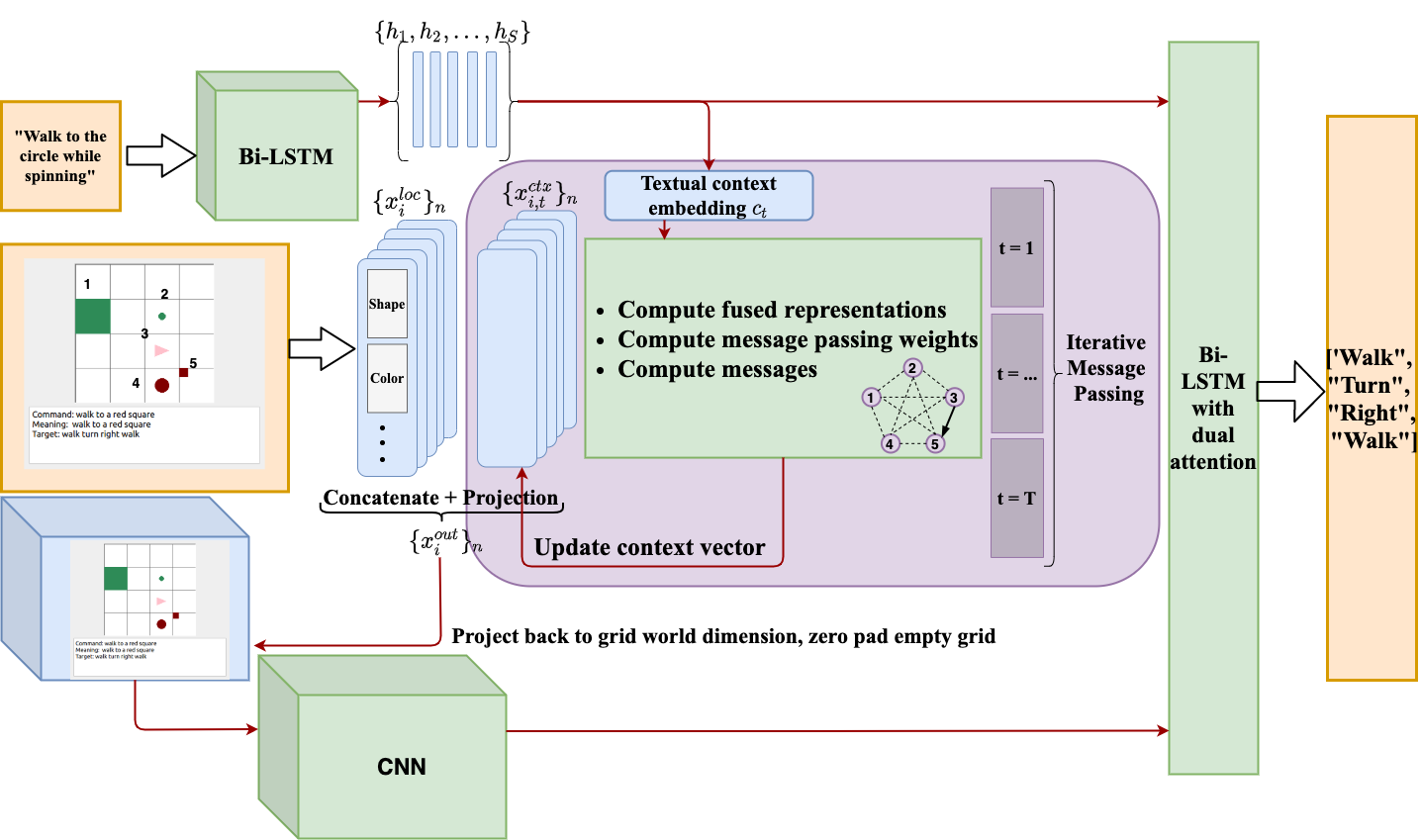}
	\caption{Model Overview}
	\label{modelfig}
\end{figure*}

\subsection{Model Architecture}
The overview of our model architecture is shown in Figure \ref{modelfig}. At the highest level, it follows the same encoder-decoder framework used by the baseline model in \citet{ruis2020benchmark} to extract information from the input sentence/grid-world representation and to output navigation instructions. However, there is a paradigm shift in how we represent and encode the grid world. Instead of viewing the grid world as a whole, we treat it as a collection of objects whose semantic meanings should be contextualized by their relations with one another. We also hypothesize that inter-object relations that are salient in a given grid world can be inferred from the accompanied natural language instruction. Therefore, we expand the vanilla CNN-based grid world encoder with a message passing module guided by the accompanied natural language instruction to obtain the contextualized grid-world embedding.

\subsubsection{Input Extraction} \label{input_extraction}

Given the input sentence and the grid world state, we first project them into higher dimensional embedding. 
For the input instruction $I=\{w_1, w_2, ..., w_S\}$ where $w_i$ is the embedding vector of word $i$,  following the practice of \citet{ruis2020benchmark} and \citet{Hu_2019}, we first encode it as the hidden states $\{h_s\}_{s=1}^S$ and the summary vector $s$ obtained by feeding the input $I$ to a Bi-LSTM as:
\begin{equation}
\label{eq:1}
    \small
    [h_1, h_2, ..., h_S] = BiLSTM(I) ~~ \text{and} ~~ s = [h_1 ; h_S]
\end{equation}
Where we use semi-colon to represent concatenation, and $h_i = [\vec{h_i};\cev{h_i}]$ is the concatenation of the forward and backward direction of the LSTM hidden state for input word $i$. For each round of message passing between the objects embedding, we further apply a transformation using a multi-step textual attention module similar to that of \citet{mac} and \citet{stacknmn} to extract the round-specific textual context. Given a round-specific projection matrix $W_2^t$, the textual attention score for word $i$ at message passing round $t$ is computed as:
\begin{equation}
    \small
    \alpha_{t, i} = \underset{s}{softmax}(W_1(h_i \odot (W_2^t ReLU (W_3s))))
\end{equation}
The final textual context embedding for message passing round $t$ is computed as:
\begin{equation}
    % \small
    c_t = \sum_{i=1}^S \alpha_{t, i} \cdot h_i\\
\end{equation}
Details of the message passing mechanism will be described in later sections.\\
As for the grid-world representation, from each grid, we extract one-hot representations of color  $\mathcal{C}$, shape $\mathcal{S}$, size $\mathcal{D}$ and agent orientation $\mathcal{O}$, and embed each property with a 16-dimensional vector. We finally concatenate them back into one vector and use this vector as the object's local embedding. 

\subsubsection{Language-conditioned Message Passing} \label{LCMP}
After extracting a textual context embedding and the objects' local embedding, we perform a language-conditioned iterative message passing for $T$ rounds to obtain the contextualized object  embeddings, where $T$ is a hyper-parameter.\\

1) Denoting the extracted object local embedding as $x^{loc}$, and previous round's object context embedding as $x^{ctx}$, we first construct a fused representation of an object $i$ at round $t$  by concatenating its local, context embedding as well their element-wise product:
\begin{equation}
\small
    x^{fuse}_{i, t} = [x^{loc}_i, x^{ctx}_{i, t-1}, (W_4 x_i^{loc}) \odot (W_5 x_{i, t-1}^{ctx})]\\
\end{equation}
We use an object's local embedding to initialize its context embedding at round 0.\\ 

2) For each pair of objects $(i, j)$, we use their fused representations, together with this round's textual context embedding to compute their message passing weight as:
\begin{equation}
\small
    w^{t}_{i, j} = \underset{s}{softmax}(W_6 x^{fuse}_{j, t})^T ((W_7 x^{fuse}_{i, t}) \odot (W_8 c_t))\\
\end{equation}
Note that the computation of the raw weight logits is asymmetric. \\

3) We consider all the objects in a grid world as nodes, and they together form a complete graph. Each node $i$ computes its message to receiver node $j$ as:
\begin{equation}
\small
    m^t_{i, j} = w^t_{i, j} \cdot ((W_9 x^{fuse}_{i, t}\odot (W_{10}c_t))\\
\end{equation}
and each receiver node $j$ updates its context embedding as:
\begin{equation}
\small
    x^{ctx}_{j, t} = W_{11}[x^{ctx}_{j, t-1}; \sum_{i=1}^{N}m^t_{i, j}]\\
\end{equation}
After $T$ rounds of iterative message passing, the final contextualized embedding for object $i$ will be:
\begin{equation}
\small
    x^{out}_i = W_{12}[x^{loc}_i; x^{ctx}_{i, T}]\\
\end{equation}

\subsubsection{Encoding the Grid World} \label{encoding_grid_world}
After obtaining contextualized embeddings for all objects in a grid world $x^s$ as $\{x^{out}\}_n = \{x^{out}_1, x^{out}_2, ..., x^{out}_n\}$ each of dimensionality $\mathcal{R}^{out}$, we map them back to their locations in the grid world, and construct a new grid world representation $X^{s'} \in \mathcal{R}^{d \times d \times out}$ by zero-padding cells without any object. This is then fed into three parallel single convolutional layers with different kernel sizes to obtain a grid world's embedding at multiple scales, as done by \citet{wang2019modeling}. The final grid world encoding is as follows:
\begin{equation}
\label{eq:9}
\small
    H^s = [H_1^s; H_2^s; H_3^s] ~~ , ~~ H_i^s = Conv_i(X^{s'})\\
\end{equation}
where ${Conv}_i$ denotes the $i$th convolutional network, and $H^s\in\mathcal{R}^{d^2\times hid}$.

\subsubsection{Decoding Action Sequences}

We use a Bi-LSTM with multi-modal attention to both the grid world embedding and the input instruction embedding to decode the final action sequence, following the baseline model provided by \citet{ruis2020benchmark}. At each step $i$, the hidden state of the decoder $h^d_i$ is computed as:
\begin{equation} 
\small
    h^d_i = LSTM([e^d_i;c^c_i;c^s_i], h^d_{i-1})
\end{equation}
where $e^d_i$ is the embedding of the previous output action token $y_{i-1}$, $c^c_i$ is the instruction context computed with attention over textual encoder's hidden states $[h^c_1, h^c_2, ..., h^c_S]$, and $c^s_i$ is the grid world context computed with attention over all locations in the grid world embedding $H^S$. We set the decoder's hidden size to 64 so that it aligns with the textual encoder, and use the attention implementation proposed by \citet{attention}. The instruction context is computed as:
\begin{align}
\small
    e^c_{ij} &= v^T_c tanh W_c(h^d_{i-1}+h^c_j)\\
    \alpha^c_{ij} &= \frac{exp(e^c_{ij})}{\sum_{j=1}^S exp(e^c_{ij})}\\
    c^c_i &= \sum_{j=1}^S \alpha^c_{ij}h^c_j, \forall j \in \{1, 2, ..., S\}
\end{align}

Similarly, the grid world context is computed as:

\begin{align}
\small
    e^s_{ij} &= v^T_s tanh W_s(h^d_{i-1}+c^c_i)\\
    \alpha^s_{ij} &= \frac{exp(e^s_{ij})}{\sum_{j=1}^{d^2} exp(e^s_{ij})}\\
    c^s_i& = \sum_{j=1}^{d^2} \alpha^s_{ij}h^s_j, \forall j \in \{1, 2, ..., d^2\}
\end{align}
where $v_s$, $v_c$, $W_c$, $W_s$ are learnable parameters, and $h^s_j$ is the embedding of grid $j$ obtained from $H^S$.\\
The distribution of next action token can then be computed as $p(y_i | x, y_1, y_2, ..., y_{i-1}) = softmax(W_oh^d_i)$.

\section{Experimental Evaluation}

\begin{table*}
\small
\centering
\begin{tabular}{|l|l|}
\hline \textbf{Split} & \textbf{Description}\\\hline
A: Random & Randomly split test sets\\\hline
B: Novel Direction & Target object is to the South-West of the agent\\\hline
C: Relativity & Target object is a size $2$ circle, referred to with the small modifier\\\hline
D: Red Squares & Red squares are the target object\\\hline
E: Yellow Squares & Yellow squares are referred to with a color and a shape at least\\\hline
F: Adverb to Verb & All examples with the adverb 'while spinning' and the verb 'pull'\\\hline
G: Class Inference & All examples where the agent needs to push a square of size 3\\\hline
\end{tabular}
\caption{\label{t1} Description of test splits}
\end{table*}

\begin{table*}
\small
\centering
\begin{tabular}{|l|l|l|l|l|}
\hline \textbf{Split} & \textbf{Baseline} & \citet{kuo2020} &\citet{Deml2020} & \textbf{Ours}\\\hline
A: Random & $97.69 \pm 0.22$ & $97.32$ & $94.19 \pm 0.71$ & $\textbf{98.6} \pm 0.95$\\\hline
B: Novel Direction & $0 \pm 0$ & $\textbf{5.73}$ & N/A & $0.16 \pm 0.12$\\\hline
C: Relativity & $35.02 \pm 2.35$ & $75.19$ & $43.43 \pm 7.0$ & $\textbf{87.32} \pm 27.38$\\\hline
D: Red Squares & $23.51 \pm 21.82$ & $80.16$ & $\textbf{81.07} \pm 10.12$ & $80.31 \pm 24.51$ \\\hline
E: Yellow Squares & $54.96 \pm 39.39$ & $95.35$ & $86.45 \pm 6.28$ & $\textbf{99.08} \pm 0.69$ \\\hline
F: Adverb to Verb    & $22.7 \pm 4.59$ & $0$ & N/A & $\textbf{33.6} \pm 20.81$\\\hline
G: Class Inference & $92.95 \pm 6.75$ & $98.63$ & N/A & $\textbf{99.33} \pm 0.46$\\\hline
\end{tabular}
\caption{\label{t2} Exact match accuracy of test splits}
\end{table*}

\subsection{Methodology \& Implementation}
We run experiments to test the hypothesis that contextualized embeddings help systematic generalization\footnote{Code is available \href{https://github.com/HQ01/gSCAN_with_language_conditioned_embedding}{here}}. Since this task has a limited vocabulary size, word-level accuracy is no longer a proper metric to reflect the model’s performance. We follow the baseline and use the exact match percentage as our metric, where an exact match means that the produced action token sequence is exactly the same as the target sequence. We compare our model with the baseline on different test sets, and use early stopping based on the exact match score on the validation set. We set the learning rate as 1e-4, decaying by 0.9 every 20,000 steps. We choose the number of message passing iterations to be 4. Our model is trained for 6 separate runs, and the average performance as well as the standard deviation are reported. Our encoder/decoder model is implemented in PyTorch \citep{pytorch} and the message passing graph network is backed by DGL \citep{dgl}. For comparison, we use test set, validation set, and baseline model released by \citet{ruis2020benchmark}.

\subsection{Results}
Table \ref{t1} is an overview of 7 test splits used for evaluation, and table \ref{t2} shows our experiment results as well as other models' performance for comparison. In the following sections, we present the results on each systematic generalization test split, and also introduce the configuration of test splits. Note that test \textbf{split A}  is a random split set that has no systematic difference from the training set. 

\textbf{Split B:} This tests the model's ability to generalize to navigation in a novel direction. For example, a testing example would require the agent to move to a target object that is to its south-west, even though during training target objects are never placed south-west of the agent. Although our model manages to predict some correct action sequences compared to the baseline's complete failure, our model still fails on the majority of cases. We further analyze the failure on Split B in the discussion section.

\textbf{Split C, G:} Split C tests the model's ability to generalize to novel contextual references. In the training set, a circle of size 2 is never referred to as ``the small circle", while in the test set the agent needs to generalize the notion ``small" to it based on its size comparison with other circles in the grid world. The message passing mechanism helps the model comprehend the relative sizes of objects, and boost the performance on split C. Besides, our model shows promising results on exploring the interrelationship between an agent and other objects in the scene, as well as learning abstract concepts by contextual comparison as shown in split G. This test split asks the model to push a square of size 3. An object with the size of 3 or 4 is defined as ``heavy", according to the configuration, and requires two consecutive push/pull actions applied on it before it actually moves. The challenge here is that the model has been trained to``pull" heavy squares and ``push"  squares with size of 4, but was never trained to ``push" a size-3 square. Thus, it needs to generalize the concept of ``heavy" and act accordingly. 

\textbf{Split D, E:} Split D and E are similar, as they both define the target object with novel combinations of color and shape. Split E is generally easier because the target object, a yellow square, appears as the target in training examples, but is only referred to as ``the square", ``the smaller square", or ``the bigger square". Split D increases the difficulty by referring to the red square, which never appears in the training set as a target but does appear as a background object. We find that while the baseline model understands the concept of ``square", it gets confused by target objects with a new color-shape combination.  In contrast, our model can generalize to novel compositions of object properties and correctly find the target object, performing significantly better on these two splits.

\textbf{Split F:} This split is designed to test the model's ability to generalize to novel adverb-verb combinations, where the model is tested under different situations but always with the terms ``while spinning" and ``pull" in the commands. However, they never appear in the training set together, consequently the model needs to generalize to this novel combination of adverb and verb. The results shows that our model does a bit better than the baseline, but  suffers from high variance across different runs.

Comparing to the two concurrent works \citet{kuo2020} and \citet{Deml2020}, our model yields better performance in general. Notice that \citet{Deml2020} and our model also report the standard deviation of multiple runs, while \citet{kuo2020} does not.

\begin{figure}[h]
\centering
\includegraphics[width=0.4\textwidth]{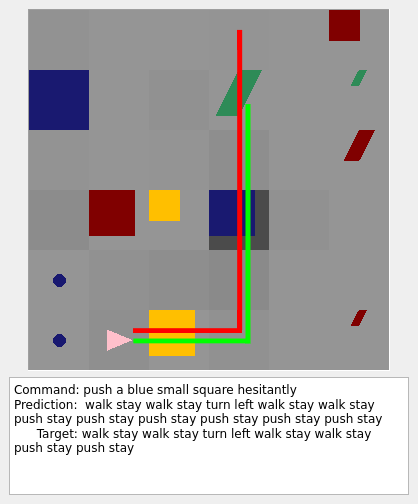}
\caption{While the target is correctly chosen, the baseline did not stop pushing even after encountering an obstacle.}
\label{fig1}
\end{figure}

\subsection{Discussion}

\begin{figure}[h]
\centering
\includegraphics[width=0.4\textwidth]{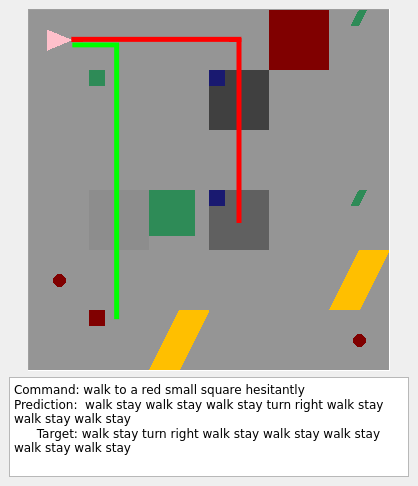}
\caption{Baseline cannot distinguish the correct square from similar candidates.}
\label{fig2}
\end{figure}

\textbf{Model Comparison.} We reveal the strength of our model by analyzing two test examples where it succeeds and the baseline fails. For each example, we visualize the grid world that the agent is in, where each cell is colored with different grey-scale levels indicating its assigned attention score. For reader's convenience, we also visualize the model's prediction and the target sequence by the red path and green path, respectively.

Figure \ref{fig1} from split G visualizes the prediction sequence as well as the attention weights generated by the baseline. The baseline attends to the position of the target object but is unable to capture the dynamic relationship between the target object and the green cylinder. It tries to push the target object over it, while our model correctly predicts the incoming collision and stops at the right time. 

Another example from split D where our model outperforms the baseline is shown in Figure \ref{fig2}. The baseline model incorrectly attends to two small blue squares and picks one as the target rather than the correct small red square. Note that the model has seen blue and green squares as targets in the training set, but has never seen a red square. This is a common mistake since the baseline struggles to choose target objects with novel property combinations when there are similar objects in the scene that were seen during training. On the contrary, our model handles these cases well, demonstrating its ability to generalize to novel color-shape combinations with the help of contextualized object embedding.

\begin{table}[h]
% \small
\centering
\begin{tabular}{|l|l|l|}
\hline \textbf{Split} & \textbf{No Message Passing} & \textbf{Full Model}\\\hline
A & $91.07 \pm 0.61$ & $98.6 \pm 0.95$\\\hline
B & $0.16 \pm 0.04$ & $0.16 \pm 0.12$\\\hline
C & $50.26 \pm 5.9$ & $87.32 \pm 27.38$\\\hline
D & $35.95 \pm 13.13$ & $80.31 \pm 24.51$ \\\hline
E & $44.18 \pm 24.56$ & $99.08 \pm 0.69$\\\hline
F & $44.82 \pm 1.95$ & $33.6 \pm 20.81$\\\hline
G & $93.02 \pm 0.33$ & $99.33 \pm 0.46$\\\hline
\end{tabular}
\caption{\label{ablation} Ablation study}
\end{table}

\textbf{Ablation Study.} We conduct an ablation study to test the significance of the language-conditioned message passing component in our network. We built a model whose architecture and hyper-parameters are the same as our full model, except that we remove the language-conditioned message passing module described in section \ref{LCMP}. That is, we follow all the steps in section \ref{input_extraction} and obtain every object's local embedding, then map new embedding back to the their locations as stated in section \ref{encoding_grid_world}. The results in Table \ref{ablation} indicate that language-conditioned message passing does help achieve higher exact match accuracy in many test splits, though it sometimes hurts the performance on split F. We conclude that the model is getting better at understanding object-related commands (``pull" moves the object), sacrificing some ability to discover the meaning of easy-to-translate adverbs that are irrelevant to the interaction with objects (``while spinning" only describes the behavior of agent with no impact on the scene).

\begin{figure}[h]
\centering
\includegraphics[width=0.4\textwidth]{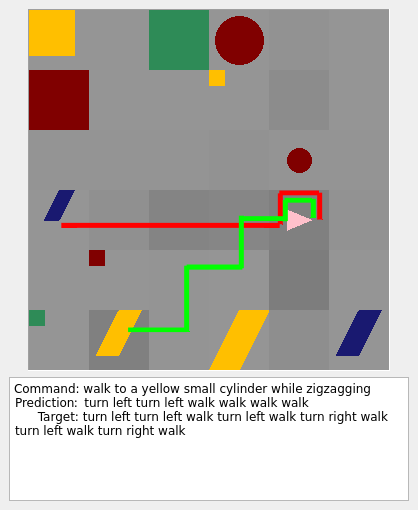}
\caption{Failure case on split B, prediction and attention scores were generated by our model.}
\label{fig3}
\end{figure}

\textbf{Failure on Split B.} Here we analyze a failure case to understand why split B is notably difficult for our model. Figure \ref{fig3} demonstrates an example that leads to both models' failure. The attention scores indicate that the model has identified the correct target position, but does not know the correct action sequence to get there. The LSTM decoder cannot generalize the meaning of action tokens that direct the agent towards an unseen direction. We can observe from our model's output prediction that, even if it manages to correctly predict the first few steps ("turn left turn left walk"), it quickly gets lost and fails to navigate to the target location. The model only observes the initial world state and the command, then generates a sequence of actions toward the target. In other words, it is blindly generating the action sequence with only a static image of the agent and the target's location, not really modeling the movement of the agent. However, humans usually do not handle navigation in a novel direction in this way. Instead, they will first turn to the correct direction, and transform the novel task into a familiar task ("walk southwest is equivalent to turn southwest then walk the same as you walk north"). This naturally requires a change of perspective and conditioning on the agent's previous action. A possible improvement is to introduce clues to inform the model of possible changes in its view as it takes actions. 

\section{Conclusion and Future Work}
In this paper, we proposed a language-conditioned message passing model for a grounded language navigation task that can dynamically extract contextualized embeddings based on input command sentences, and can be trained end-to-end with the downstream action-sequence decoder. We showed that obtaining such contextualized embeddings improves performance on a recently introduced challenge problem, gSCAN, significantly outperforming the state-of-the-art across several test splits designed to test a model's ability to represent novel concept compositions and achieve systematic generalization. 

Nonetheless, our model's fairly poor performance on split B and F shows that challenges still remain. As explained in the discussion section, our model is falling short of estimating the effect of each action on the agent's state. An alternative view of this problem is as a reinforcement learning task with sparse reward. Sample-efficient model-based reinforcement learning  \citep{buckman2018sample} could then be used, and its natural ability to explicitly model  environment change should improve performance on this task.

It would also be beneficial to visualize the dynamically generated edge weights during message passing to have a more intuitive understanding of what contextual information is integrated during the message passing phase. Currently, we consider all objects appearing on the grid, including the agent, as homogeneous nodes during message passing, and all edges in the message passing graph are modelled in the same way. However, intuitively, we should model the relation between different types of objects differently. For example, the relation between the agent and the target object of pulling might be different from the relation between two objects on the grid. Inspired by \citet{systematic}, it would be interesting to try modeling different edge types explicitly with neural modules, and perform type-specific message passing to obtain better contextualized embeddings.

\bibliographystyle{acl_natbib}
\bibliography{aacl-ijcnlp2020}

\newpage
\appendix
\section{Appendix}
\subsection{Implementation Details}

Our implementation is based on the gSCAN dataset used by the \citet{ruis2020benchmark} and the world size is $d=6$.

For equation \ref{eq:1}, each  token is embedded to a randomly initialized vector of size 32, and the hidden size of the encoder BiLSTM is 32. 

For equation \ref{eq:9}, we use three convolutional networks with kernel size $k=1, 5, 7$ and padding size $\lfloor \frac{k}{2} \rfloor=0, 2, 3$ to ensure that the resulting dimensionality is synchronized with input. They share the same filter size of 64. The concatenation of $H^s_i$ is also flattened to the shape of $36\times 192$.

Table \ref{tp} presents the shapes of other trainable parameters mentioned in section \ref{sec:3}. We simply set $d_{cmd}=d_h=d_{loc}=d_m=d_s=d_{ctx}=64$.

\begin{table}[h]
% \small
\centering
\begin{tabular}{|l|l|}
\hline \textbf{Parameter} & \textbf{Shape} \\\hline
$W_1$ & $1\times d_{cmd}$ \\\hline
$W_2, W_3$ & $d_{cmd}\times d_{cmd}$ \\\hline
$W_4$ & $d_h\times d_{loc}$ \\\hline
$W_5$ & $d_h\times d_{ctx}$ \\\hline
$W_6, W_7$ & $d_h\times (d_{loc}+d_{ctx}+d_h)$ \\\hline
$W_8$ & $d_h\times d_s$ \\\hline
$W_9$ & $d_m\times (d_{loc}+d_{ctx}+d_h)$ \\\hline
$W_{10}$ & $d_m\times d_s$ \\\hline
$W_{11}$ & $d_{ctx}\times d_{ctx}$ \\\hline
$W_{12}$& $d_h\times (d_{loc}+d_{ctx})$ \\\hline
\end{tabular}
\caption{\label{tp} Parameter Shapes}
\end{table}
\subsection{Example Visualization}
\label{sec:appendix}
\begin{figure}[h] 
\centering
\includegraphics[width=0.4\textwidth]{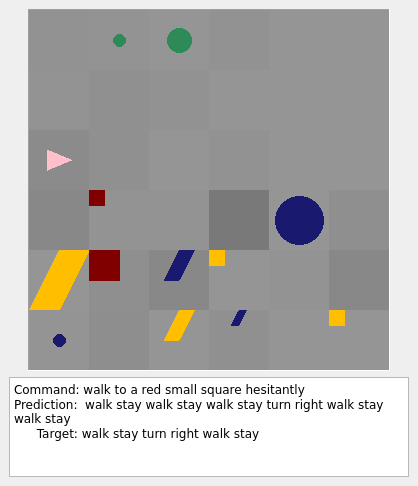}
\caption{Baseline incorrectly picked a yellow square as the target.} \label{figb1}
\end{figure}

\begin{figure}[h]
\centering
\includegraphics[width=0.4\textwidth]{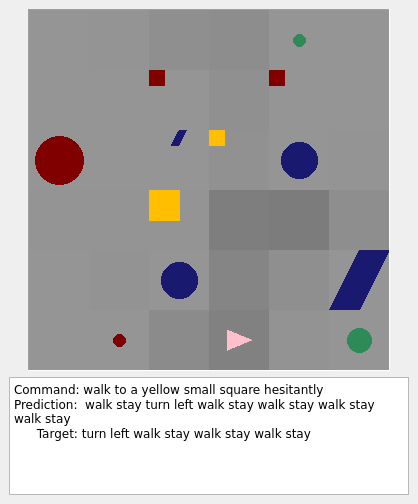}
\caption{Baseline incorrectly picked a red square as the target.} \label{figb2}
\end{figure}

\begin{figure}[h]
\centering
\includegraphics[width=0.4\textwidth]{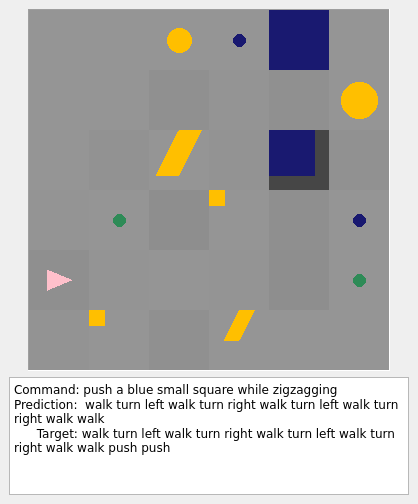}
\caption{Baseline falsely predicted the consequential interaction and decided not to push.} \label{figb3}
\end{figure}

\begin{figure}[h]
\centering
\includegraphics[width=0.4\textwidth]{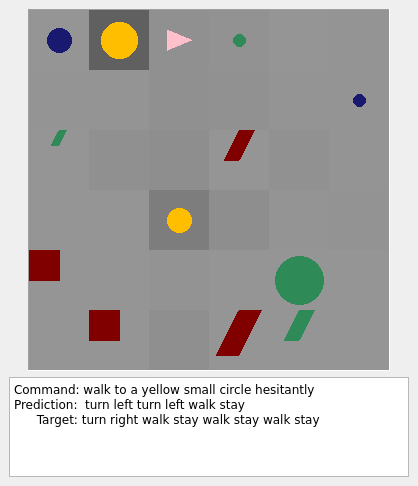}
\caption{Baseline incorrectly picked the bigger circle instead of the smaller one.} \label{figb4}
\end{figure}

\begin{figure}[h]
\centering
\includegraphics[width=0.4\textwidth]{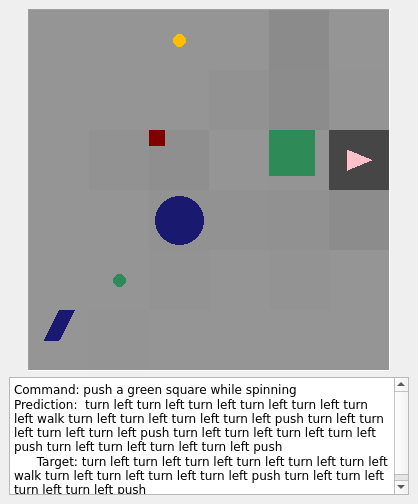}
\caption{Getting lost along a long sequence: Our model fails when the target sequence repeats the same actions several times.} \label{figm1}
\end{figure}

\begin{figure}[h]
\centering
\includegraphics[width=0.4\textwidth]{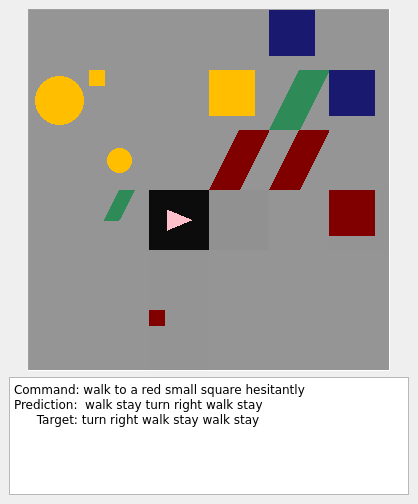}
\caption{Incorrect path plan: Our model generates the path plan in a partially-reversed order.} \label{figm2}
\end{figure}

\begin{figure}[h]
\centering
\includegraphics[width=0.4\textwidth]{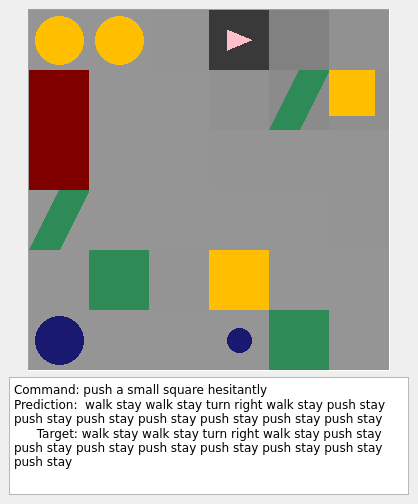}
\caption{Early stop before reaching boundary: Our model stops pushing when the target object is next to the boundary grid.} 
\label{figm3}

\end{figure}

Here we present more examples demonstrating our model's strengths and weaknesses. 
Figures \ref{figb1} - \ref{figb4} are cases where our model's prediction exactly matches the target while the baseline's does not. Some of the common failures of our model are illustrated in Figures \ref{figm1} - \ref{figm3}.



\end{document}